# Feature-Fused SSD: Fast Detection for Small Objects

Guimei Cao, Xuemei Xie, Wenzhe Yang, Quan Liao,Guangming Shi, Jinjian Wu
School of Electronic Engineering, Xidian University, China
xmxie@mail.xidian.edu.cn

## ABSTRACT

Small objects detection is a challenging task in computer vision due to its limited resolution and information. In order to solve this problem, the majority of existing methods sacrifice speed for improvement in accuracy. In this paper, we aim to detect small objects at a fast speed, using the best object detector Single Shot Multibox Detector (SSD) with respect to accuracy-vs-speed trade-off as base architecture. We propose a multi-level feature fusion method for introducing contextual information in SSD, in order to improve the accuracy for small objects. In detailed fusion operation, we design two feature fusion modules, concatenation module and element-sum module, different in the way of adding contextual information. Experimental results show that these two fusion modules obtain higher mAP on PASCAL VOC2007 than baseline SSD by 1.6 and 1.7 points respectively, especially with 2-3 points improvement on some small objects categories. The testing speed of them is 43 and 40 FPS respectively, superior to the state of the art Deconvolutional single shot detector (DSSD) by 29.4 and 26.4 FPS. Code is available at *https://github.com/wnzhyee/ Feature-Fused-SSD*.

## 1. INTRODUCTION

Reliably detecting small objects is a quite challenging task due to their limited resolution and information in images. Considering this problem, many existing methods [1-4] have demonstrated considerable improvement brought by exploiting the *contextual* information. For example, in Fig. 1, it is quite difficult to recognize the sailing boats without taking the sea into account. Besides, both the people and bike provide an evidence for the existing of the bottle and the person, which have a relationship in some extent. Another common method used for small object detection is *enlarging* the small regions [5] for better fitting the features of pre-trained network. Since the enlarging method would increase the computation greatly, we do not consider it.

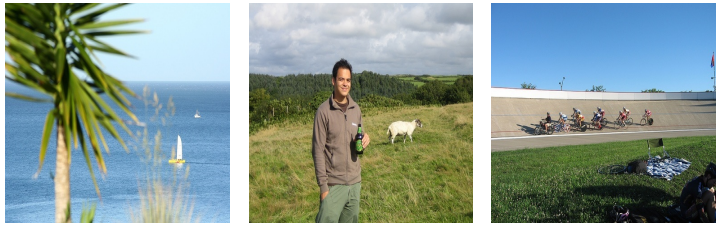

Figure 1. The scene of sea, person, and bike provide the evidence of the existence of the sailing boats, bottle, and person.

When considering real-time detection, most of these studies are based on region-based object detection architecture, including RCNN [6], SPPnet [7], Fast RCNN [8], Faster RCNN [9], which cannot detect small objects in a fast way. The fast detector Single Shot Multibox Detector (SSD [10]) presents large improvement in speed, for eliminating region proposals and the subsequent pixel resampling stage. To improve accuracy for small objects, Deconvolutional Single Shot Detector (DSSD [4]) uses the SSD architecture as baseline. Since it is concentrated on improving accuracy by using the base network residual-101 [11], it sacrifices lot of speed inevitably. Thus, for small object detection, accuracy-vs-speed trade-off is still hard to balance currently, remaining a challenging problem to be solved.

In this paper, we aim to obtain fast detection for small objects. To achieve this, we use the best object detector SSD with respect to accuracy-vs-speed trade-off as our base architecture, which can do real-time detection. We propose a multi-level feature fusion method for adding contextual information in SSD baseline, in order to improve the accuracy for small objects. In detailed fusion operation, we instantiate this feature fusion method carefully with two modules, concatenation module and element-sum module. Since context may introduce useless background noises, it is not always useful for small object detection. In detail, concatenation module uses a 1×1 convolution layer for learning the weights of the fusion of the target information and contextual information, which can reduce the interference of useless background noises. Element-sum module uses equivalent weights set manually and fuses the multi-level features in a compulsory way, which can enhance the effectiveness of useful context. Experimental results show that our two feature fusion modules obtain higher mAP on PASCAL VOC2007 than baseline SSD by 1.6 and 1.7 points respectively, especially with 2-3 points improvement on some small objects categories, such as aero, bird, boat, pot plant, TV monitor and so on. The testing speed of them is 43 and 40 FPS respectively, which is in a real-time fashion.

## 2. RELATED WORK

**Context:** Many previous studies have demonstrated that contextual information plays an important role in object detection task, especially for small objects. The common method for introducing contextual information is exploiting the combined feature maps within a ConvNet for prediction. For example, ION [2] extracts VGG16 [12] features from multiple layers of each region proposal using ROI pooling [8], and concatenate them as a fixed-size descriptor for final prediction. HyperNet [3], GBD-Net [13] and AC-CNN [14] also adopt a similar method that use the combined feature descriptor of each region proposal for object detection. Because the combined features come from different layers, they have different levels of abstraction of input image. So that the feature descriptors of each region proposal contain fine-grained local features and contextual features. However, these methods are all based on region proposal method and pool the feature descriptors from combined feature maps, which increases the memory footprint as well as decreases the speed of detection.

**Multi-scale representation:** Multi-scale representation has been proven useful for many detection tasks. Many previous detection architectures use *single-scale* representation, such as RCNN [6], Fast RCNN [8], Faster RCNN [9], and YOLO [15]. They predict confidence and localization from the features extracted by the top-most layer within a ConvNet, which increases the heavy burden of the last layer. Differently, SSD [10] uses *multi-scale* representation that detect objects with different scales and aspect ratios from multiple layers. With smaller object detection, SSD uses the features from the shallower layers, while exploits the features from the deeper layers for bigger objects detection. In order to further improve the accuracy of SSD, especially for small objects, DSSD adds extra deconvolution layers in the end of SSD. By integrating every prediction layer and its deconvolution layer, the contextual information is injected, which makes the predictions for small objects more accurate.

Since the methods for introducing contextual information into region-based object detectors are also multi-scale representation, they are quite different from SSD in the way of predicting objects localization and confidence.

## 3. MULTI-LEVEL FEATURE FUSION MODULE IN SSD

We extract the multi-level features from SSD model, which is the state of the art object detector with respect to accuracy-vs-speed trade-off. In this section, we introduce the SSD briefly and then illustrate the two multi-level feature fusion modules in detail, which exploit the useful local context for final classification and regression.

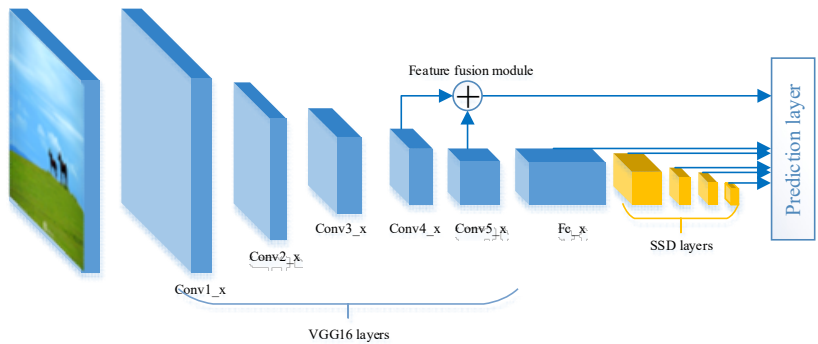

Figure 2. Feature-fused SSD architecture.

## 3.1 Single Shot Multibox Detector

The Single shot multibox detector (SSD) is based on the VGG16 base network and ends with several newly added layers. Our model is illustrated in Fig. 2. Instead of using the last feature map of the ConvNet, SSD uses the pyramidal feature hierarchy in multiple layers within a ConvNet to predict objects with different scales. That is using shallower layers to predict smaller objects, while using deeper layers to predict bigger objects, thus it can reduce the prediction burden of the whole model. However, shallower layers often lack of semantic information which is important supplement for small object detection. Therefore, passing the semantic information captured in convolutional forward computation back to the shallower layers will improve the detection performance of small objects.

**Which layers to combine?** We exploit the proper Conv layers to provide useful contextual information since that too large receptive field would often introduce large useless background noise. Because of SSD predicting small objects with its shallower layers, such as conv4_3, we do not use the feature fusion module for large objects in the deeper layers for less decrease of speed. For choosing the proper feature fusion layers, effective receptive fields in different layers are explored with deconvolution method. As shown in Fig. 3, take boats for example, the effective receptive field of the nearer boat within the SSD architecture is quite proper in conv4_3. Conv5_3 and fc6 have larger effective receptive fields. However, fc6, where uses dilated convolution, would introduce more background noise than conv5_3.

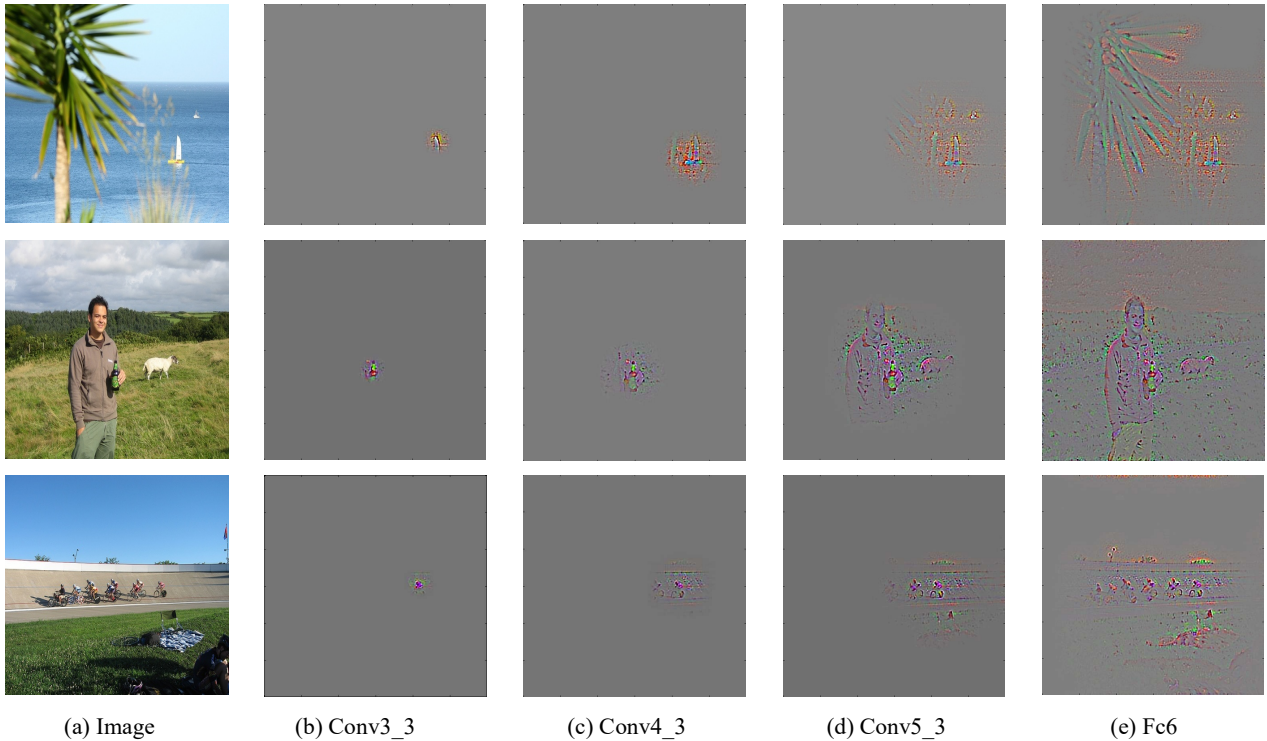

(a) Image    (b) Conv3_3    (c) Conv4_3    (d) Conv5_3    (e) Fc6

Figure 3. Effective receptive fields in SSD architecture. (a) is the input image, and (b), (c), (d), (e) show the effective fields of conv3_3, conv4_3, conv5_3 and fc6, respectively, shown with enhancement.

In order to inject the contextual information into the shallower layers (conv4_3) which lacks of sematic information, we design two different modules for feature fusion, named as concatenation module and element-sum module.

## 3.2 Concatenation Module

The concatenation fusion module is shown in Fig. 4. In order to make the feature maps of conv5_3 layer the same size as conv4_3 layer, the conv5_3 layer is followed by a deconvolution layer, which is initialized by bilinear upsample. Two 3×3 convolutional layers are used after conv4_3 layer and conv5_3 layer for learning the better features to fuse. Then normalization layers are following with different scales respectively, i.e. 10, 20, before concatenating them along their channel axis. The final fusion feature maps are generated by a 1×1 convolutional layer for dimension reduction as well as feature recombination.

### 3.3 Element-Sum Module

Another solution to fuse feature maps of two layers is using element-sum module. This module is illustrated in Fig. 5. It is the same as concatenation module except the fusion type. In this module, two feature maps contained different level features are summarized point to point with equivalent weights. In practice, this operation works well due to the two convolution layers used before, which learns features from conv4_3 and conv5_3 adaptively for better fusion effects. This module is inspired by DSSD [4] based on residual-101, which uses the learned deconvolution layer and element-wise operation.

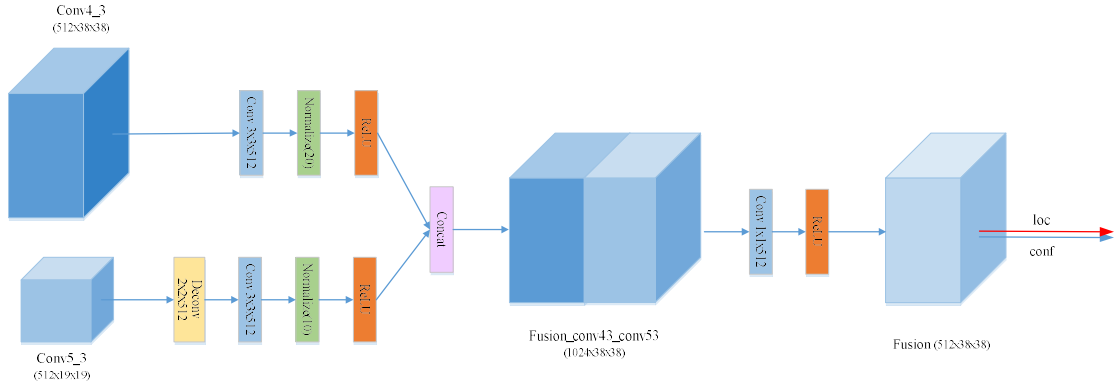

Figure 4. Illustration of the concatenation module.

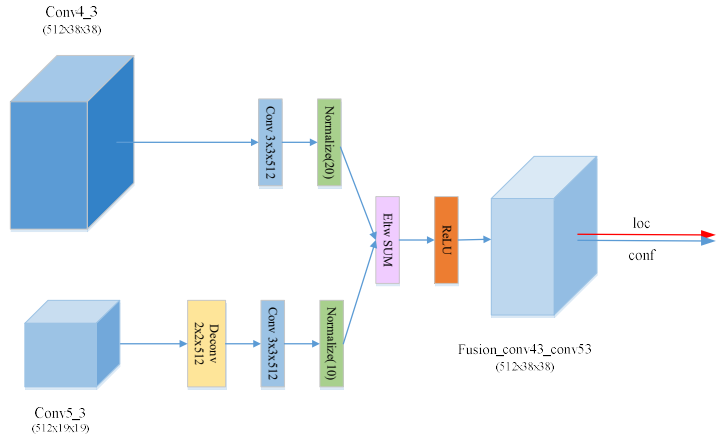

Figure 5. Illustration of the element-sum module.

Concatenation module fuses multi-level features with learning weights implemented by 1×1 convolutional layer, while element-sum module uses the equivalent weights set manually. With this difference, concatenation module can reduce the interference caused by useless background noises, and element-sum module can enhance the importance of the contextual information.

## 4. EXPERIMENTAL RESULTS

We evaluate our feature-fused SSD model on PASCAL VOC 2007 and 2012. In this section, the detection results of two feature fusion modules are compared with SSD and DSSD. In addition, the detection performance for small objects is shown and the comparisons are analyzed carefully. And we illustrate the testing speed of our two models in the end.

### 4.1 Experimental Setup

The feature-fused SSD is implemented based on SSD built on the Caffe [16] framework, and the VGG16 architecture, all of which are available online. The VGG16 model pre-trained on ImageNet dataset for the task of image classification, is reduced of fully connected layers when used as the base network of SSD. The baseline SSD is trained with the batch size of 16, and with the input size of 300×300. The training process starts with the learning rate at $10^{-3}$ for the first 80K iterations, which decreases to $10^{-4}$ and $10^{-5}$ at 10K and 12K.

## 4.2 PASCAL VOC2007

We train our models on the union of PASCAL VOC2007 and PASCAL VOC2012, which contain 20 categories in 9,963 and 22,531 images, respectively. Both of the two feature fusion models are fine-tuned upon the well-trained SSD baseline for another 10K iterations. The learning rate is $10^{-3}$ for the first 60K iterations and then decreases to $10^{-4}$ and $10^{-5}$ at 60K and 70K iterations respectively. All newly added layers initialized with "xavier".

We take several necessary tries when designing the most effective feature fusion modules. Firstly, we explore the proper layers to fuse with experimental results, which has been discussed in Figure 3 theoretically. As shown in Table 1, the mAP on general objects of PASCAL VOC 2007 is considerable. But we select the conv4_3 and conv5_3 layers to fuse, since fc6 has a larger receptive field than conv5_3 for small objects which would introduce much more background noises. In addition, we try different numbers of kernels when designing the modules. In Table 2, we show that our module can work well even if the number of kernels reducing to 128, which means smaller and faster.

Table 1. Detection results of different fusion layers

| layers | concat | Eltsum |
|---|---|---|
| Conv4_3 | 77.27 | |
| Conv4_3+conv5_3 | **78.76** | 78.53 |
| Conv4_3+fc6 | 78.56 | 78.51 |
| Conv3_3+Conv4_3+conv5_3 | 78.48 | 78.39 |

Table 2. Detection results of different number of kernels

| Kernel number | concat | Eltsum |
|---|---|---|
| 512 | **78.76** | 78.53 |
| 384 | 78.63 | **78.92** |
| 256 | 78.70 | 78.83 |
| 128 | 78.68 | 78.67 |
| 64 | 78.44 | 78.37 |
| 32 | 78.05 | 78.14 |

**General objects detection:** In Table 3, we report the detection performance of the proposed modules. The two feature-fused SSD methods are both improved compared with their baseline SSD with respect to general objects detection. The feature-fused SSD with concatenation module obtains 78.8 mAP, while 78.9 mAP with element-sum module, which are 1.6 and 1.7 points higher than original SSD respectively. Moreover, our results are comparable with the state of the art performance DSSD 321.

Table 3. Results on PASCAL VOC2007 test set (with IOU=0.5)

| Method | mAP | Aero | bike | bird | boat | bottle | bus | car | cat | chair | cow |
|---|---|---|---|---|---|---|---|---|---|---|---|
| SSD300 | 77.2 | 78.8 | 85.3 | 75.7 | 71.5 | 49.1 | 85.7 | 86.4 | 87.8 | 60.6 | 82.7 |
| DSSD321 | 78.6 | 81.9 | 84.9 | **80.5** | 68.4 | **53.9** | 85.6 | 86.2 | **88.9** | 61.1 | **83.5** |
| Elt_sum | **78.9** | 82.0 | **86.5** | 78.0 | 71.7 | 52.9 | 86.6 | 86.9 | 88.3 | **63.2** | 83.0 |
| Concat | 78.8 | **82.4** | 85.7 | 77.8 | **73.8** | 52.3 | **87.5** | 86.8 | 87.6 | 62.6 | 82.1 |
| Method | mAP | table | dog | horse | mbike | person | plant | sheep | sofa | train | tv |
| SSD300 | 77.2 | 76.5 | 84.9 | 86.7 | 84.0 | 79.2 | 51.3 | 77.5 | 78.7 | 86.7 | 76.3 |
| DSSD321 | 78.6 | **78.7** | **86.7** | **88.7** | 86.7 | 79.7 | 51.7 | 78.0 | **80.9** | 87.2 | **79.4** |
| Elt_sum | **78.9** | 76.8 | 86.1 | 88.5 | **87.5** | **80.4** | **53.9** | **80.6** | 79.5 | **88.2** | 77.9 |
| Concat | 78.8 | 76.6 | 86.1 | 88.2 | 86.6 | 80.3 | 53.7 | 78.0 | 80.1 | 87.3 | 78.0 |

**Small objects detection:** Since VOC2007 dataset contains 20 categories and every category may have small objects, we manually select 181 images mainly including small objects for evaluating our models better. These two methods achieve 3.6 and 2.0 mAP improvement compared with original SSD model, with respect to concatenation module and element-sum module. Detection results are shown in Fig. 6. We find that the detection performance of small objects with specific background improved obviously, such as small car, aero, bird and so on. And objects which often appear together with relative objects are detected more accurate. For example, the people in a car as well as the bike beside a person provide supplementary evidence for the existence of each other.

**Performance comparison of two fusion modules:** Consider *small object* detection. When looking into the difference of these two feature-fused SSD methods, we analyze fusion methods and their detection results carefully. In Fig. 7, a pot plant is occluded by a desk bump, which is not the useful contextual information of it. Since the concatenation module uses learned weights to combine the target feature and the contextual feature, it can select useful contextual information and weaken the interference of the background noises. Unfortunately, element-sum model combines both target features and context in an equivalent way, thus it cannot select the useful contextual information adaptively. On the contrary, in Fig. 8, cars are blurry in the scene, so that the context is necessary for detection. In this case, element-sum module works better than the concatenation module, because the latter has much more choice that may not learn the relationship between the target and context well.

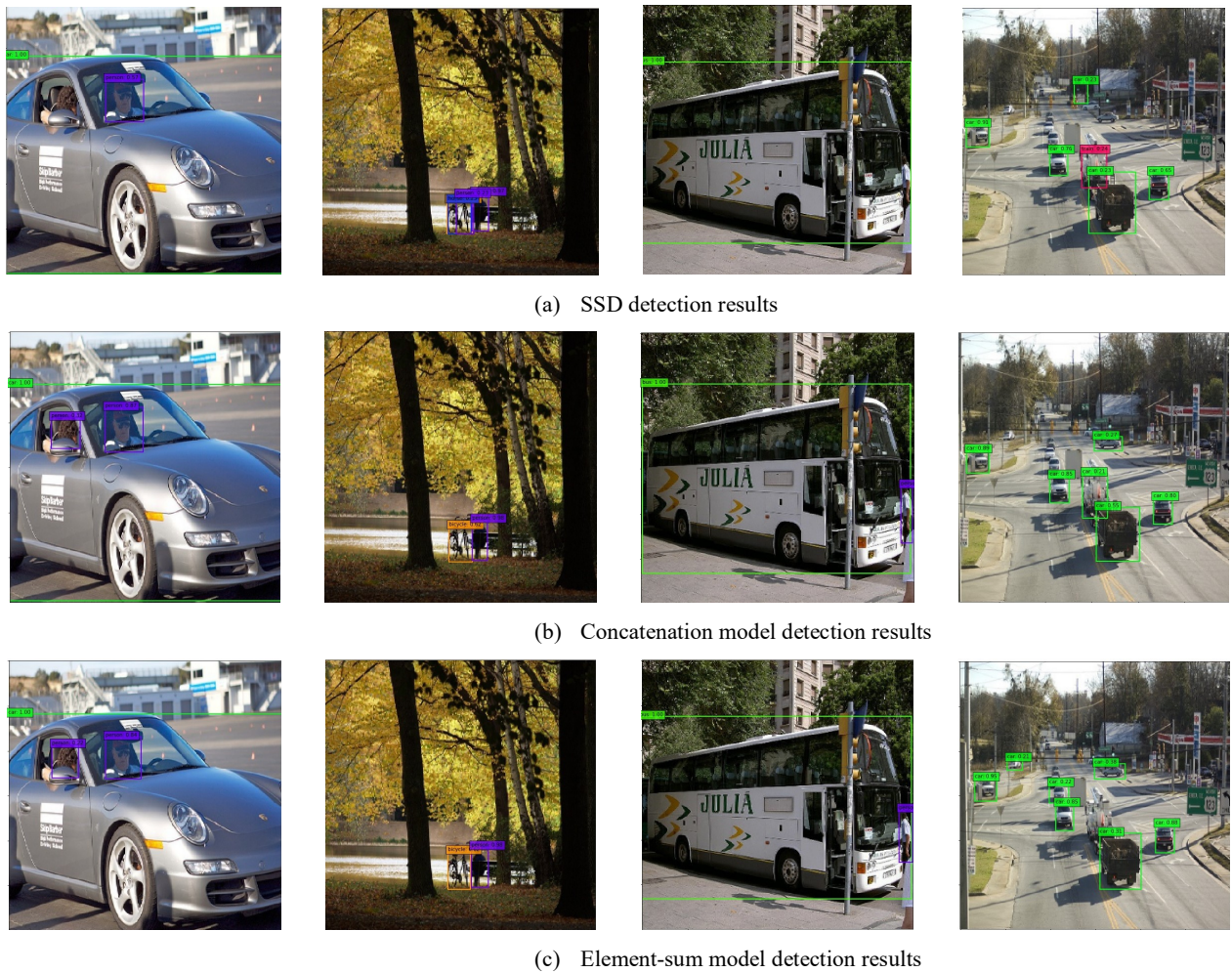

(a) SSD detection results

(b) Concatenation model detection results

(c) Element-sum model detection results

Figure 6. Detection results of feature-fused SSD. (a), (b), and (c) are detection results of original SSD, feature-fused SSD with concatenation module, and with element-sum module, respectively. All the images in four columns show that useful contextual information provide evidence for the existence of the small objects.

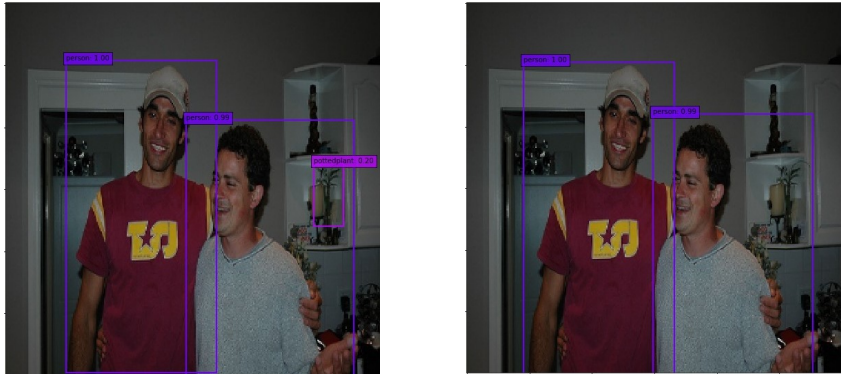

Figure 7. Left: Detection results of concatenation model. Right: Detection results of element-sum model. The pot plant is occluded by a desk bump, and concatenation model can weaken the interference of background noises, while element-sum model cannot.

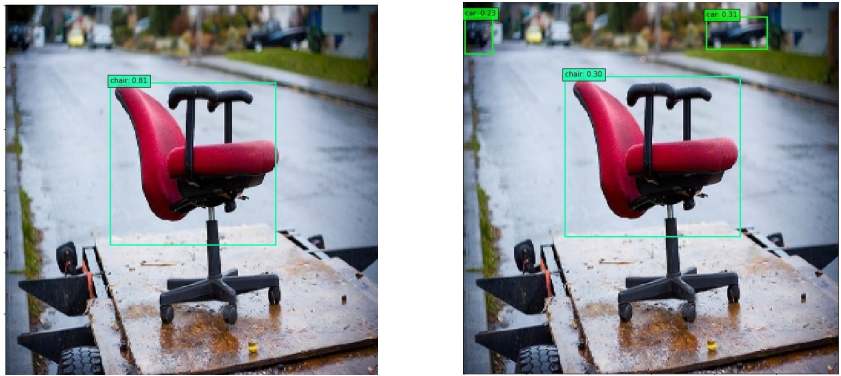

Figure 8. Left: Detection results of concatenation model. Right: Detection results of element-sum model. The cars in this image is small and blurry, so that the contextual information is necessary for their detection. Element-sum model exploits this context quite well, while concatenation model cannot.

### 4.3 Running Time

We evaluate running time for these two feature fusion methods on PASCAL VOC 2007 test dataset, as shown in Table 4. The detection speed of the two methods, concatenation module and element-sum module, is 40 FPS and 43 FPS respectively, slower than SSD original model because of the extra feature fusion layers. However, our methods still achieve real-time detection. Compared with the DSSD321 with 13.6 FPS, faster detection speed is obtained by our methods, with comparable accuracy 78.8 and 78.9. Because DSSD321 uses Residual-101 network as base network, while our methods exploit VGG16.The concatenation model uses three convolution layers with 512 kernels in each layer, and element-sum module uses two convolution layers with 384 kernels in each layer. Thus, the element-sum model is faster than the concatenation model by 3 FPS.

Table 4. The running time illustration of different models

| Method | Base Network | mAP | FPS |
| --- | --- | --- | --- |
| SSD300[10] | VGG16[12] | 77.2 | 50 |
| DSSD321[4] | Residual-101[11] | 78.6 | 13.6 |
| Proposed concatenation model | VGG16 | 78.8 | 40 |
| Proposed element-sum model | VGG16 | **78.9** | **43** |

## 5. CONCLUSION

We have presented a feature fusion method in SSD obtains a fast and accurate detection for small objects. Compared with the state of the art object detector for small objects, our method achieves faster detection speed, with comparable accuracy. With the two fusion operations, we have shown the advantages of them in different cases. Since context is not always useful information for small object detection, which may introduce useless background noises sometimes, controlling the information transmission will be the further work to study.

## ACKNOWLEDGMENT

This work is supported by Natural Science Foundation (NSF) of China (61472301, 61632019).

## REFERENCES

[1] C. Chen, M. Liu, O. Tuzel, J. Xiao, "R-cnn for small object detection," Asian Conference on Computer Vision, 214-230, (2016)

[2] S. Bell, C. Lawrence Zitnick, K. Bala, R. Girshick, "Inside-outside net: Detecting objects in context with skip pooling and recurrent neural networks," Proceedings of the IEEE Conference on Computer Vision and Pattern Recognition, 2874-2883, (2016)

[3] T. Kong, A. Yao, Y. Chen, F. Sun, "HyperNet: Towards accurate region proposal generation and joint object detection," Proceedings of the IEEE Conference on Computer Vision and Pattern Recognition, (2016)

[4] C. Fu, W. Liu, A. Ranga, A. Tyagi, A. Berg, "DSSD: Deconvolutional single shot detector," arXiv preprint arXiv:1701.06659, (2017)

[5] P. Hu, D. Ramanan, "Finding tiny faces," arXiv preprint arXiv:1612.04402, (2016)

[6] R. Girshick, J. Donahue, T. Darrell, J. Malik, "Rich feature hierarchies for accurate object detection and semantic segmentation," Proceedings of the IEEE conference on computer vision and pattern recognition, 580-587, (2014)

[7] K. He, X. Zhang, S. Ren, J. Sun, "Spatial pyramid pooling in deep convolutional networks for visual recognition," European Conference on Computer Vision, 346-361, (2014)

[8] R. Girshick, "Fast r-cnn," Proceedings of the IEEE international conference on computer vision, 1440-1448, (2015)

[9] S. Ren, K. He, R. Girshick, J. Sun, "Faster R-CNN: Towards real-time object detection with region proposal networks," Advances in neural information processing systems, 91-99, (2015)

[10] W. Liu, D. Anguelov, D. Erhan, C. Szegedy, S. Reed, C. Fu, A. Berg, "SSD: Single shot multibox detector," European conference on computer vision, 21-37, (2016)

[11] K. He, X. Zhang, S. Ren, J. Sun, "Deep residual learning for image recognition," Proceedings of the IEEE conference on computer vision and pattern recognition, 770-778, (2016)

[12] K. Simonyan, A. Zisserman, "Very deep convolutional networks for large-scale image recognition," arXiv preprint arXiv:1409.1556, (2014).

[13] X. Zeng, W. Ouyang, J. Yan, H. Li, T. Xiao, K. Wang, Y. Liu, Y. Zhou, B. Yang, Z. Wang, H. Zhou, X. Wang, "Crafting GBD-Net for Object Detection," arXiv preprint arXiv:1610.02579, (2016)

[14] J. Li, Y. Wei, X. Liang, J. Dong, T. Xu, J. Feng, S. Yan, "Attentive contexts for object detection," IEEE Transactions on Multimedia, **19**(5): 944-954, (2017)

[15] J. Redmon, S. Divvala, R. Girshick, A. Farhadi, "You only look once: Unified, real-time object detection," Proceedings of the IEEE Conference on Computer Vision and Pattern Recognition, 779-788, (2016)

[16] Y. Jia, E. Shelhamer, J. Donahue, S. Karayev, J. Long, R. Girshick, S. Guadarrama, T. Darrell, "Caffe: Convolutional architecture for fast feature embedding," Proceedings of the 22nd ACM international conference on Multimedia, 675-678, (2014)